\documentclass[10pt,twocolumn,letterpaper]{article}

\usepackage{cvpr}
\usepackage{times}
\usepackage{epsfig}
\usepackage{graphicx}
\usepackage{amsmath}
\usepackage{amssymb}
\usepackage{float}
\usepackage{subfigure}
\usepackage{algorithm}
\usepackage{algorithmic}

\usepackage{bm}
\usepackage{color}
\usepackage{booktabs}
\usepackage{multirow}

\usepackage[pagebackref=true,breaklinks=true,letterpaper=true,colorlinks,bookmarks=false]{hyperref}

\cvprfinalcopy 

\def\cvprPaperID{76} 

\ifcvprfinal\fi
\begin{document}

\title{Learning a Discriminative Null Space for Person Re-identification}
\author{Li Zhang \quad \quad Tao Xiang \quad \quad Shaogang Gong\\
Queen Mary University of London\\
{\tt\small \{david.lizhang, t.xiang, s.gong\}@qmul.ac.uk}
}

\maketitle

\begin{abstract}
Most existing person re-identification (re-id) methods focus on learning the optimal distance metrics across camera views. Typically a person's appearance is represented using features of thousands of dimensions, whilst only hundreds of training samples are available due to the difficulties in collecting matched training images. With the number of training samples much smaller than the feature dimension, the existing methods thus face the classic small sample size (SSS) problem and have to resort to dimensionality reduction techniques and/or matrix regularisation, which lead to loss of discriminative power. In this work, we propose to overcome the SSS problem in re-id distance metric learning by matching people in a discriminative  null space of the training data. In this null space, images of the same person are collapsed into a single point thus minimising the within-class scatter to the extreme and maximising the relative between-class separation simultaneously. Importantly, it has a fixed dimension, a closed-form solution and is very efficient to compute. Extensive experiments carried out on five person re-identification benchmarks including VIPeR, PRID2011, CUHK01, CUHK03 and Market1501 show that such a simple approach beats the state-of-the-art alternatives, often by a big margin.
\end{abstract}

\section{Introduction}
The problem of person re-identification (re-id) has attracted great attention in the past five years \cite{Vezzan_survey,Gong_book_re-id}. When a person is captured by multiple non-overlapping views, the objective is to match him/her across views among a large number of imposters. Despite the best efforts from the computer vision researchers, re-id remains a largely unsolved problem. This is because that a person's appearance often undergoes dramatic changes across camera views due to changes in view angle, body pose, illumination and background clutter. Furthermore, since people are mainly distinguishable by their clothing under a surveillance setting, many passers-by can be easily confused with the target person because they wear similar clothes.


		\begin{figure}[H]
	\centering
		\includegraphics[height=6cm]{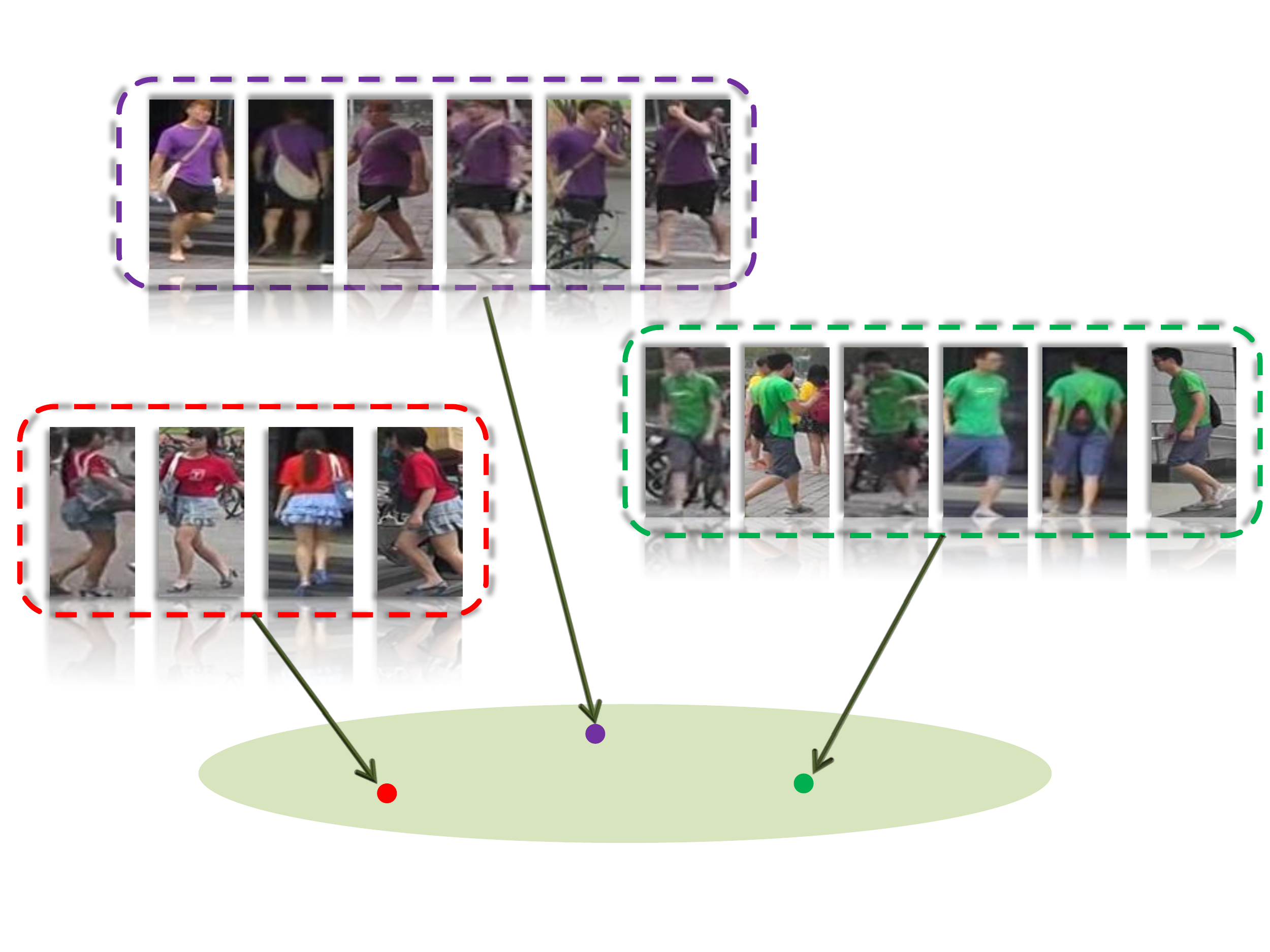}
		\label{fig:null space}
		\vspace{-0.3cm}
		\caption{Training images of same identity are projected to a single point in a learned discriminative null space. 
		}
	\end{figure}

Existing approaches focus on developing discriminative feature representations that are robust against the view/pose/illumination/background changes \cite{Gray_PedestrianRecognitionEnsembleLocalFeature_08a,yang2014salient,Farenzena_cvpr10,
  kviatkovsky2013color,ma2012local,zhao2014learning,liao2015person}, or learning a  distance metric  \cite{Gray_PedestrianRecognitionEnsembleLocalFeature_08a,koestinger2012large,
  Prosser_PerReSVR_10a,Zheng_RDC_2013a,mignon2012pcca,li2013person,pedagadi2013local,
  li2013learning,zhao2013unsupervised,zhao2013person,xiong2014person,ma2014person,lisantiperson,liao2015person}, or  both jointly \cite{li2014deepreid,ahmed2015improved}.  Among them, the distance metric learning methods are most popular and are the focus of this paper. Given any feature representation and a set of training data consisting of matching image pairs across camera views, the objective is to learn the optimal distance metric that gives small values to images of the same person and large values for those of different people. Distance metric learning has been extensively studied in machine learning \cite{yang2006distance}, and existing
metric learning methods employed for re-id are either originated elsewhere or extensions of existing methods with modifications to address the additional challenges arising from the re-id task. Although they have been shown to be effective in improving the existing re-id benchmarks over the past five years, all these models are still limited by some of classical problems in model learning.

Specifically, a key challenge for distance metric learning when applied to person re-id is the small sample size (SSS) problem \cite{ChenLKLY00}. Specifically, to capture rich person appearance whilst being robust against those condition changes mentioned above, the feature representations used by most recent re-id works are of high dimension -- typically in the order of thousands or tens of thousands. In contrast, the number of training samples is typically small, normally in hundreds. This is because that collecting training samples of matched person pairs across views is labour intensive and tedious. As a result the sample size is much smaller (often in an order of magnitude) than the feature dimension, a problem known as the SSS problem. Metric learning methods suffer from the SSS problem because they essentially aim to minimise the within-class (intra-person) variance (distance), whilst maximising the inter-class (inter-person) variance (distance). With a small sample size, the within-class scatter matrix becomes singular \cite{ChenLKLY00}; to avoid it,  unsupervised dimensionality reduction or regularisation are required. This in turn makes the learned distance metric sub-optimal and less discriminative \cite{ChenLKLY00,ZhengZZ05,guo2006null}. 


In this paper, we argue that the SSS problem in person re-id distance metric learning can be best solved by learning a discriminative null space of the training data. In particular, instead of minimising the within-class variance, data points of the same classes are {\em collapsed}, by a transform, into a single point in a new space (see Fig.~\ref{fig:null space}). By keeping the between-class variance non-zero, this automatically maximises the Fisher discriminative criterion and results in a discriminative subspace.  The null space method, also known as the null Foley-Sammon transfer (NFST) \cite{guo2006null} is specifically designed for the small sample case, with rigorous theoretical proof on the resulting subspace dimension. Importantly, it has a closed-form solution, no parameter to tune, requires no pre-precessing steps to reduce the feature dimension, and can be computed efficiently.  Furthermore, to deal with the non-linearity of the person's appearance, a kernel version can be developed easily to further boost the matching performance within the null space. It therefore offers a perfect solution to the challenging person re-id problem. In addition to formulating the NSFT model as a fully supervised model to solve the person re-id problem, we also extend it to the semi-supervised setting to further alleviate the effects of the SSS problem by exploiting unlabelled data abundant in re-id applications.  

The contributions of this work are as follows: (1) We identify the small sample size (SSS) problem suffered by all existing metric learning based re-id methods and argue that their solutions to this problem is suboptimal. (2) For the first time, we propose to overcome the SSS problem in person re-id by learning a discriminative null space of the training data. (3) We  develop a novel semi-supervised learning method in the null space to exploit the abundant unlabelled data to further alleviate the effects of the SSS problem. Extensive experiments carried out on five person re-identification benchmarks including VIPeR \cite{gray2007evaluating}, PRID2011 \cite{hirzer2011person}, CUHK01 \cite{CUHK_dataset}, CUHK03 \cite{CUHK_dataset} and Market1501 \cite{zheng2015scalable} show that such a simple and computationally very efficient approach beats all state-of-the-art methods presented to date, often by a large margin.

\section{Related Work}


Existing works on person re-id can be roughly categorised into three groups.
The first group of methods design invariant and discriminant features
\cite{li2014deepreid,Gray_PedestrianRecognitionEnsembleLocalFeature_08a,Farenzena_cvpr10,
  kviatkovsky2013color,ma2012local,zhao2014learning,lisanti2014matching,yang2014salient,liao2015person}.  The general trend is that the dimensions of the proposed features are getting higher. For instance the dimensions of two representations, recently proposed  in \cite{lisanti2014matching} and \cite{liao2015person} and used in our experiments, are 5,138 and 26,960 respectively. However, no matter how robust the designed features are, they are unlikely to be completely invariant to the often drastic cross-view pose/illumination/background changes. Therefore, the second group of methods
focus on learning robust and discriminative distance metrics or subspaces for matching people across views
\cite{Gray_PedestrianRecognitionEnsembleLocalFeature_08a,koestinger2012large,
  Prosser_PerReSVR_10a,Zheng_RDC_2013a,mignon2012pcca,li2013person,pedagadi2013local,
  li2013learning,zhao2013unsupervised,zhao2013person,xiong2014person,ma2014person,lisanti2014matching,lisantiperson,liao2015person,paisitkriangkrailearning}. Recently, the third group of methods start to  appear which are based on deep learning \cite{li2014deepreid,ahmed2015improved}. However, person re-id seems to be one of the few vision problems that deep learning has not been able to shine due to the small training sample size problem addressed in this work.


Apart from a few exceptions \cite{Gray_PedestrianRecognitionEnsembleLocalFeature_08a,Prosser_PerReSVR_10a} based on ranking or boosting, the second groups of methods can be further divided into two major sub-groups: those on learning distance metrics \cite{koestinger2012large,Zheng_RDC_2013a,mignon2012pcca,li2013person} and those on learning discriminative subspaces \cite{pedagadi2013local,lisanti2014matching, xiong2014person,liao2015person}.  Seemingly different, these two sub-groups are closely related \cite{MCML_GlobersonR05}. Specifically, most metric learning methods focus on Mahalanobis form metrics. If the linear projection of a feature vector $\mathbf{x}_i$ in a learned discriminative subspace is denoted as $\mathbf{y}_i$, we have $\mathbf{y}_{i} = \mathbf{W}^{T}\mathbf{x}_{i}$. The Euclidean distance between $\mathbf{y}_{i}$ and $\mathbf{y}_{j}$ is exactly a Mahalanobis distance $||\mathbf{y}_{i}-\mathbf{y}_{j}||^2 =(\mathbf{x}_{i} - \mathbf{x}_{j})^T \mathbf{A} (\mathbf{x}_{i} - \mathbf{x}_{j})$ where $\mathbf{A} = \mathbf{W}^{T}\mathbf{W}$ is a positive semidefinite matrix. In other words, learning a discriminative subspace followed by computing Euclidean distance is equivalent to computing a discriminative Mahalanobis distance over feature vectors in the original space. By making this connection, it is not difficult to see why both methods suffer from the same SSS problem typically associated with the subspace learning methods \cite{ChenLKLY00,ZhengZZ05,guo2006null}. Most existing methods need to work with a reduced dimensionality \cite{pedagadi2013local}, achieved typically by PCA whose dimension has to be carefully tuned for each dataset. Some works additionally require introducing matrix regularisation term if the intra-class scatter matrix is used in the formulation, in order to prevent matrix singularity \cite{pedagadi2013local,lisanti2014matching, xiong2014person,liao2015person}, again with free parameters to tune. Critically, they suffer from  the degenerate eigenvalue problem (i.e.~several eigenvectors share the same eigenvalue), which makes the solution sub-optimal resulting in  loss of discriminant ability \cite{ZhengZZ05}. In contrast, for our discriminative  null space based approach, neither dimensionality reduction before model learning nor regularisation term is required,   and it has no parameters to tune. 



As a solution proposed specifically to address the SSS problem, the null Foley-Sammon transfer (NFST) method has been around for a long time \cite{guo2006null}, but received very little attention apart from a recent application to the novelty detection problem \cite{bodesheim2013kernel}. A possible reason is that by restricting the learned discriminative projecting directions to the null projecting directions (NPDs), on which within-class distance is always zero and between-class distance is positive, the model is {\em extreme}, leaving little space for further extension with clear added-value. For example, the more relaxed Fisher discriminative analysis (FDA) can be extended, gaining notable advantage, by exploit graph laplacian to preserve local data structure, known as LFDA\cite{sugiyama2006local}, which has been successfully applied to re-id \cite{pedagadi2013local}. However, a similar graph laplacian extension to NFST does not apply due to its single point per class nature. Despite the restrictions,  given the acute SSS problem in re-id distance metric learning, the basic idea of learning a null space for overcoming this problem becomes very attractive. The general concept of collapsing same-class data points to a single point has been exploited in a Mahalanobis distance learning framework, known as maximally collapsing metric learning (MCML) \cite{MCML_GlobersonR05}. However, MCML does not exploit a null space. Instead, the MCML model must make approximations with plenty of free parameters to tune and no
closed-form solution.


In this work, we exploit the original null Foley-Sammon transfer (NFST) method
\cite{guo2006null} with its conventional supervised
learning approach, benefiting from its attractive closed-form
solution and no parameters tuning required. Moreover, we extend the
original fully supervised null space model to a semi-supervised 
learning setting. This is to explore, in addition to a few labelled data, a larger
quantities of unlabelled data typically available in person re-id
scenarios for model learning. The problem of semi-supervised re-id has
attracted interest lately due to its potential to overcome the lack of
training data problem. One approach is by dictionary
learning for sparse coding \cite{liu2014semi,kodirovdictionary}, which
has an unsupervised nature, thus  can be learned with
both labelled and unlabelled data. In this work, we compare the new
semi-supervised null space model against dictionary learning based
methods and demonstrate the superior performance from the new model.

\section{Methodology}
\subsection{Problem Definition}

Given a set of $N$ training data denoted as $\mathbf{X} \in \Bbb{R}^{d \times N}$. Each column of the data descriptor matrix $\mathbf{X}$, $\mathbf{x}_i$ is a feature vector representing the $i$-th training sample.  In the case of person re-id, this feature vector is extracted from a person detection box and contains appearance information about the person, and its dimension $d$ is typically very high. We assume that each data point belongs to one of  $C$ classes, i.e.~$C$ different identities. The objective of learning a discriminative null space is to learn a projection matrix $\mathbf{W} \in \Bbb{R}^{d \times m}$ to project the original high-dimensional feature vector $\mathbf{x}_i$ into a lower-dimensional one $\mathbf{y}_i \in \Bbb{R}^{m}$ with $m<d$. Person re-id can then be performed by computing the Euclidean distance between two projected vectors in the learned discriminative null space. 

\subsection{Foley-Sammon Transform}
The learned null  Foley-Sammon transform (NFST) space is closely related to linear discriminant analysis (LDA), also known as Foley-Sammon transform (FST) \cite{NFS_75}. So before we formulate NFST, let us first briefly revisit  FST. 

The objective of FST is to learn a projection matrix $\mathbf{W} \in \Bbb{R}^{d \times m}$ so that each column, denoted as $\mathbf{w}$, is an optimal discriminant direction that maximises the Fisher discriminant criterion:
\begin{equation}\label{fisher}
\mathcal{J}(\mathbf{w}) = \frac{\mathbf{w}^{\top}\mathbf{S}_{b} \mathbf{w} }{\mathbf{w}^{\top}\mathbf{S}_{w} \mathbf{w} },
\end{equation}
where $\mathbf{S}_{b}$ is the  between-class scatter matrix and $\mathbf{S}_{w}$ is the  within-class scatter matrix. The optimisation of Eq.~(\ref{fisher}) can be done by solving the following generalised eigen-problem:
\begin{equation} 
\mathbf{S}_{b} \mathbf{w} = \lambda \mathbf{S}_{w} \mathbf{w}.
\end{equation}

If $\mathbf{S}_{w}$ is non-singular, $C-1$ eigenvectors $\mathbf{w}^{(1)},...,\mathbf{w}^{(c-1)}$ can be computed corresponding to the $C-1$ largest eigenvalues of $\mathbf{S}_{w}^{-1}\mathbf{S}_{b}$. Using them as the columns, the projection matrix  $\mathbf{W}$ can project the original data into a $C-1$ dimensional  discriminative subspace where the $C$ classes become maximally separable. However, in the small sample size case, we have $d>N$; as a result, $\mathbf{S}_{w}$ is singular. FST thus runs in numerical problems and common solutions include reducing $d$ by PCA or adding a regularisation term to $\mathbf{S}_{w}$. In \cite{guo2006null}, a more principled way to overcome the SSS problem in FST is proposed, termed as  Null  Foley-Sammon transform (NFST).

\subsection{Null Foley-Sammon transform}

NFST aims to learn a discriminative subspace where the training data points of each of the $C$ classes are collapsed to a single point, resulting in $C$ points in the space. In order to make this subspace discriminative, these $C$ points should not further collapse to a single point. Formally, we aim to learn the optimal projection matrix $\mathbf{W}$ so that each of its column $\mathbf{w}$ satisfies the following two conditions: 
\begin{equation}\label{fisher0} 
\mathbf{w}^{\top}\mathbf{S}_{w} \mathbf{w} = 0,
\end{equation}
\begin{equation}\label{fisher1} 
\mathbf{w}^{\top}\mathbf{S}_{b} \mathbf{w} > 0.
\end{equation}
That is, it satisfies zero within-class scatter and positive between-class scatter. This guarantees the best separability of the training data in the sense of  
Fisher discriminant criterion.  Such a linear projecting direction $\mathbf{w}$ is called Null Projecting Direction (NPD) \cite{guo2006null}.

Next, we show that a NPD must lie in the null space of $\mathbf{S}_{w}$. In particular, we have the following Lemma:

\emph{\textbf{Lemma 1.} Let $\mathbf{W}$ be a projection matrix which maps a sample $\mathbf{x}$  into the null space of $\mathbf{S}_{w}$, where the null space is spanned by the orthonormal set of $\mathbf{W}$, that is,  $\mathbf{S}_{w}\mathbf{W} = 0$. If all samples are mapped into the null space of $\mathbf{S}_{w}$ through $\mathbf{W}$, the within-class scatter matrix $\widehat{\mathbf{S}}_{w}$ of the mapped samples is a complete zero matrix.}

\textbf {Proof.} Let $\mathbf{x}_{n}^{c}$ be the $n^{th}$ sample of the $c^{th} \in \left \{1,...C\right \}$ class which has $N_c$ samples in total. $\mathbf{y}_{n}^{c}$ denote the mapped feature vector through $\mathbf{W}$. We have:
\vspace{-0.2cm}
\begin{align*}
\widehat{\mathbf{S}}_{w} &= \sum^{C}_{c=1} \sum^{N_{c}}_{n=1}(\mathbf{y}_{n}^{c} - \mathbf{\overline{y}}^{c})(\mathbf{y}_{n}^{c} - \mathbf{\overline{y}}^{c})^{\top} \\
&=\sum^{C}_{c=1} \sum^{N_{c}}_{n=1}(\mathbf{W}^{\top}\mathbf{x}_{n}^{c} - \mathbf{W}^{\top}\mu^{c})(\mathbf{W}^{\top}\mathbf{x}_{n}^{c} - \mathbf{W}^{\top}\mu^{c})^{\top}\\
&= \mathbf{W}^{\top}\sum^{C}_{c=1} \sum^{N_{c}}_{n=1}(\mathbf{x}_{n}^{c} - \mu^{c})(\mathbf{x}_{n}^{c} - \mu^{c})^{\top}\mathbf{W}\\
&=\mathbf{W}^{\top}\mathbf{S}_{w}\mathbf{W} = \mathbf{0}
\end{align*}
where $\mathbf{y}_{n}^{c} = \mathbf{W}^{\top}\mathbf{x}_{n}^{c}$, $\mathbf{\overline{y}}^{c} = \mathbf{W}^{\top}\mathbf{\mu}^{c}$, $\mathbf{\mu}^{c} = \frac{1}{N_{c}} \sum^{N_{c}}_{n=1}\mathbf{x}^{c}_{n}$, $N_{c}$ is the number of samples in class $c$, and $\mu^{c}$ is the mean vector of all data belonging to the class $c$.

Now with Lemma 1, we know that  Eq.~(\ref{fisher0}) holds as long as $\mathbf{w}$ is from the null space of $\mathbf{S}_{w}$. Next we take a look the condition in the  inequality~(\ref{fisher1}). It is easy to see that when Eq.~(\ref{fisher0}) holds, (\ref{fisher1}) also holds if:
\begin{equation} 
\mathbf{w}^{\top}\mathbf{S}_{t} \mathbf{w} > 0,
\end{equation}
where $\mathbf{S}_{t} = \mathbf{S}_{b} + \mathbf{S}_{w}$ is the total scatter matrix. We now denote the null space of the $\mathbf{S}_{t}$ and $\mathbf{S}_{w}$as:
\begin{equation} 
\mathbf{Z}_{t} =  \left \{  \mathbf{z}\in \mathbb{R}^{d}\mid \mathbf{S}_{t}\mathbf{z}=0\right \},
\end{equation}
\begin{equation} 
\mathbf{Z}_{w} =  \left \{  \mathbf{z}\in \mathbb{R}^d\mid \mathbf{S}_{w}\mathbf{z}=0\right \},
\end{equation}
and their orthogonal complements as $\mathbf{Z}_{t}^{\perp}$ and $\mathbf{Z}_{w}^{\perp}$ respectively. 
Now since $\mathbf{S}_{b}$ is non-negative definite, we can see that in order for the NPDs to satisfy both Eqs.~(\ref{fisher0}) and (\ref{fisher1}) simultaneously, they must  lie in the shared space between $ \mathbf{Z}_{w}$ and $\mathbf{Z}_{t}^{\perp}$, that is: 
\begin{equation} 
\mathbf{w} \in (\mathbf{Z}_{t}^{\perp}\cap  \mathbf{Z}_{w}).
\end{equation}

It has been proved in \cite{guo2006null} that there are precisely $C-1$ NPDs $\mathbf{w}$ that satisfy both Eq.~(\ref{fisher0}) and (\ref{fisher1}). In other words, the discriminative null space we are looking for has $m=C-1$ dimensions.

\subsection{Learning the Discriminative Null Space}

Let $\mathbf{X}_{w}$ be the matrix consisting of vectors $\mathbf{x}_{i}^{c} - \mu^{c}$. $\mathbf{X}_{t}$ be the matrix consisting of vectors $\mathbf{x}_{i} - \mu$ with $\mu = \frac{1}{N}\sum^{N}_{i=1}\mathbf{x}_{i}$. We then have,

\begin{equation}\label{represen}
\mathbf{S}_{w} = \frac{1}{N} \mathbf{X}_{w} \mathbf{X}_{w}^{\top},\quad \mathbf{S}_{t} = \frac{1}{N} \mathbf{X}_{t} \mathbf{X}_{t}^{\top}
\end{equation}

Now we know where to look for the NPDs -- the shared space between $ \mathbf{Z}_{w}$ and $\mathbf{Z}_{t}^{\perp}$. Next, we shall see how to  compute them. Let us first take a look at how to compute $\mathbf{w} $ that satisfies $\mathbf{w}  \in \mathbf{Z}_{t}^{\perp}$. First we notice that:
\begin{align*}
\mathbf{Z}_{t} &=  \left \{  \mathbf{z}\in \mathbb{R}^{d}\mid \mathbf{S}_{t}\mathbf{z}=0\right \}  = \left \{\mathbf{z}\in \mathbb{R}^{d}\mid   \mathbf{z}^{\top}\mathbf{S}_{t}\mathbf{z} = 0 \right \} \\
&= \left \{\mathbf{z}\in \mathbb{R}^{d}\mid   (\mathbf{X}_{t}^{\top}\mathbf{z})^{\top}\mathbf{X}_{t}^{\top}\mathbf{z} = 0 \right \} \\
&=  \left \{  \mathbf{z}\in \mathbb{R}^{d}\mid \mathbf{X}_{t}^{\top}\mathbf{z}=0\right \}.
\end{align*} 
Hence, $\mathbf{Z}_{t}^{\perp}$ is the subspace spanned by zero-mean data $\mathbf{x}_{i} - \mu$. We can obtain the orthonormal basis $\mathbf{U} = [\mathbf{u}^{(1)},...,\mathbf{u}^{(N-1)}]$ of the zero-mean data using Gram-Schmidt orthonormalisation, then represent each solution $\mathbf{w}$ as:
\begin{equation}\label{basis} 
\mathbf{w} = \beta_{1}\mathbf{u}^{(1)}+...+\beta_{N-1}\mathbf{u}^{(N-1)}=\mathbf{U}\beta,
\end{equation}
Note that there are $N-1$ basis vectors because the rank of $\mathbf{S}_{t}$ is $N-1$.

So now after expressing $\mathbf{w}$ using Eq.~(\ref{basis}), it must satisfy $\mathbf{w}  \in \mathbf{Z}_{t}^{\perp}$. The next step is the make it also satisfy $\mathbf{w}  \in \mathbf{Z}_{w}$. This can be achieved by substituting Eq.~(\ref{basis}) into Eq.~(\ref{fisher0}) and solve the following  eigen-problem:
\begin{equation}\label{solution} 
(\mathbf{U}^{\top}\mathbf{S}_{w}\mathbf{U})\beta = \mathbf{0},
\end{equation}
for which we know that $C-1$ solutions $\beta ^{(1)},...,\beta ^{(C-1)}$ exist, giving $C-1$ NPDs, $\mathbf{U}\beta$. 

In summary, the problem of learning the discriminative null space boils down to solving an eigen-problem which has a closed-form solution and can be solved very efficiently. Importantly, the whole optimisation algorithm has no free parameter to tune.

\subsection{Kernelisation}
The NFST model is a linear model. It has been demonstrated \cite{xiong2014person} that many distance metric learning or discriminative subspace based methods for person re-id benefit from kernelisation because of the non-linearity in person's appearance. In the following we describe how the discriminative null space can be kernelised. 

Given a kernel function $k(\mathbf{x}_i,\mathbf{x}_i) = \langle \Phi(\mathbf{x}_i),\Phi(\mathbf{x}_j)\rangle$, where $\Phi(\mathbf{x}_i)$ maps $\mathbf{x}_i$ to an implicit higher dimensional space, we can compute the data kernel matrix $\mathbf{K} \in \Bbb{R}^{N \times N}$ for training data $\mathbf{X}$  as $\mathbf{K} =  \Phi(\mathbf{X})^{\top} \Phi(\mathbf{X})$.
Now the  within-class scatter matrix $\mathbf{S}_{w}$ and total-class scatter matrix $\mathbf{S}_{t}$ can be kernelised as:
\begin{align*}
&\mathbf{K}_{w} = \mathbf{K}(\mathbf{I}-\mathbf{L})(\mathbf{I}-\mathbf{L})^{\top}\mathbf{K}, \\
&\mathbf{K}_{t} = \mathbf{K}(\mathbf{I}-\mathbf{M})(\mathbf{I}-\mathbf{M})^{\top}\mathbf{K},
\end{align*}
where $\mathbf{I}$ is a $N \times N$ identity matrix, $\mathbf{L}$ is a block diagonal matrix with block sizes equal to the number of data points $N_{c}$ for each class $c \in \left \{ 1,...C \right \}$ and $\mathbf{M}$ is a $N \times N$ matrix with all entries equal to $\frac{1}{N}$.

Now to write Eq.~(\ref{solution}) in its kernelised form, we need to replace $\mathbf{S}_{w}$ with $\mathbf{K}_{w}$, and compute the orthonormal basis of $\mathbf{K}_{t} $ to replace $\mathbf{U}$. The orthonormal basis of $\mathbf{K}_{t} $ can be computed using kernel PCA. First, we compute the centred kernel matrix $\widetilde{\mathbf{K}}$. Second, the eigendecomposition of $\widetilde{\mathbf{K}}$ is written as  $\mathbf{K}_{t} = \mathbf{V}\mathbf{E}\mathbf{V}^{\top} $ with $\mathbf{E}$ being
the diagonal matrix containing $N-1$ non-zero eigenvalues
and $\mathbf{V}$ containing the corresponding eigenvectors in its
columns. Now the scaled eigenvectors $\widetilde{\mathbf{V}} = \mathbf{V}\mathbf{E}^{-1/2}$ contain
coefficients for the kernelised orthonormal basis used to replace  $\mathbf{U}$ in  Eq.~(\ref{solution}). 
Let 
\begin{equation} 
\mathbf{H} = ((\mathbf{I} - \mathbf{M})\widetilde{\mathbf{V}})^{\top}\mathbf{K}(\mathbf{I}-\mathbf{L}),
\end{equation}
and with Eq.~(\ref{represen}), we can rewrite Eq.~(\ref{solution})
as:
\begin{equation}\label{product}
\mathbf{H}\mathbf{H}^{\top}\beta = \mathbf{0}.
\end{equation}
By solving the eigen-problem  Eq.~(\ref{product}), we obtain the final $C-1$ null projection directions (NPDs) as:
\begin{equation}\label{finalprojection} 
\mathbf{w}^{(i)} = ((\mathbf{I} - \mathbf{M})\widetilde{\mathbf{V}})^{\top}\beta ^{(i)} \quad \forall i=1,...,C-1.
\end{equation}

\subsection{Semi-supervised Learning}

The NFST method is a fully supervised method. When applied to the problem of re-id, the labelled training set is used to learn the projection $\mathbf{W}$. The test data are then projected into the same subspace and matched by computing the Euclidean distance between a query sample and a set of gallery samples. 

In a real-world application scenario, the labelled training data are scarce but there are often plenty of unlabelled data (person images collected from different views) that can be used to alleviate the small sample size problem. To this end, the   NFST method is extended to the semi-supervised setting. More specifically, given a training set $\mathbf{X}$ contains a labelled subset $\mathbf{X}^l$ of $N^l$ samples  and an unlabelled subset $\mathbf{X}^u$ of $N^u$ samples. Using the NFST method described above, we can first learn an initial projection matrix $\mathbf{W}^0$ using $\mathbf{X}^l$ only. Then $\mathbf{X}^u$ is projected to the lower-dimensional subspace through $\mathbf{W}^0$ and becomes $\mathbf{Y}^u_{\mathbf{W}^0}$. To utilise the unlabelled data $\mathbf{X}^u$, we use their projections $\mathbf{Y}^u_{\mathbf{W}^0}$ to build a cross-view correspondence matrix $\mathbf{A} \in
\Bbb{R}^{N_u \times N_u}$ which
captures the identity relationship for the unlabelled people across views. Note, since the  data are unlabelled, the
true cross-view correspondence relationship is unknown. We therefore
use $\mathbf{A}$ to represent a soft cross-view correspondence relationship. That
is, each person in one view can correspond to multiple people in another view depending
on their visual similarity in the learned discriminative subspace parameterised by $\mathbf{W}^0$.   To this end, we first
construct a $k$-nearest-neighbour ($k$-nn) graph $G$ across camera views with $N_u$
vertices, where each vertex represents a unlabelled data point.  $\mathbf{A}$  is then
computed as the weight matrix of $G$ using a heat kernel. With this $k$-nn graph, we then create pseudo-classes, each consisting one vertex from one view and its $k$-nearest-neighbours from the other view. Next these pseudo-classes are augmented with the labelled classes in $\mathbf{X}^l$ to create a new training set, denoted $\mathbf{P}$, on which a new project matrix  $\mathbf{W}^1$ is computed using NFST. 
Re-learning the projection matrix runs iteratively till the average distance for the $k$-nearest-neighbours stop decreasing. In our experiments, we found that the algorithm converges rapidly. 

This semi-supervised learning is essentially based on self-training, a popular strategy taken by many semi-supervised learning methods \cite{zhu2005semi}. For any self-training based methods, preventing model drift is of paramount importance. Apart from  examining the average distance for the $k$-nearest-neighbours, another measure taken is to rank the $k$-nearest-neighbours and take only the top $f$ percent with the smallest distance to create the pseudo-classes. The complete semi-supervised null space learning algorithm is summarised in Alg. \ref{alg1}.

\renewcommand{\algorithmicrequire}{\textbf{Input:}}
\renewcommand{\algorithmicensure}{\textbf{Output:}}

\begin{algorithm}[!t]
\caption{Semi-supervised null space learning}
\label{alg1}
	\begin{algorithmic}[1]
		\REQUIRE $\mathbf{X}^l$, $\mathbf{X}^u$, $k$, $\mathbf{P}^0 = \mathbf{0}$.
		\ENSURE The learned projection $\mathbf{W}$.\\
		\STATE Estimate $\mathbf{W}^0$ using $\mathbf{X}^l$;
		\STATE $t=0$;
			\WHILE{\textit{not converged}} 
				\STATE project $\mathbf{X}^u$ through $\mathbf{W}^t$ to obtain $\mathbf{Y}^u_{\mathbf{W}^t}$
				\STATE build $k$-nn graph $G$ with $\mathbf{Y}^u_{\mathbf{W}^t}$
				\STATE take top $f$ percent to create the pseudo-classes $\mathbf{P}^{t+1}$
\STATE learn $\mathbf{W}^{t+1}$ with $\mathbf{X}^l+\mathbf{P}^{t+1}$
				 \STATE $t = t+1$
			\ENDWHILE

	\end{algorithmic}
\end{algorithm}

\section{Experiments}


\subsection{Datasets and Settings}

\noindent{\bf Datasets} \quad Five widely used datasets are selected for experiments, including the three largest benchmarks available (CUHK01, CUHK03, and Market1501).

\noindent \textbf{VIPeR} ~\cite{gray2007evaluating}  contains 632 identities and each has two images captured
outdoor
from two views with distinct view angles.
All images are scaled to 128 $\times$ 48 pixels. The 632 people's images are randomly divided into two equal halves, one for training and the other for testing. This is repeated for 10 times and the averaged performance is reported. 

\noindent{\bf PRID2011}~\cite{hirzer2011person} consists of person images recorded from two cameras. 
Specifically,  it has two camera views.  View $A$ captures  385 people, whilst View $B$ contains 749 people. Only 200 people appear in both views. The single shot version of the dataset is used in our experiments as in \cite{hirzer2012relaxed}: In each data split, 100 people with one image from each view are randomly chosen from the 200 present in both camera views for the training set, while the remaining 100 of View $A$ are used as the probe set, and the remaining 649 of View $B$ are used as gallery. Experiments are repeated over the 10 splits provided in \cite{hirzer2012relaxed}.

\noindent{\bf CUHK01} \cite{CUHK_dataset}  
contains 971 identities with each person having two images in each camera view. All the images are normalised to 160 $\times$ 60 pixels. Following the standard setting, images from camera A are used as probe and those from camera B as gallery. We randomly partition the dataset into 485 people for training and 486 for testing (multi-shot) following \cite{liao2015person, zhao2014learning}, again over 10 trials.

\noindent{\bf CUHK03} \cite{li2014deepreid}  contains 13,164 images of 1,360 identities, captured by six surveillance cameras with each person only appearing in two views. It provides both manually labelled pedestrian bounding boxes and bounding boxes automatically detected by the deformable-part-model (DPM) detector~\cite{felzenszwalb2010object}. 
A real-world re-id system has to rely on a person detector; the latter version of the data is thus ideal for testing performance given detector errors.  
We report results on both of the manually labelled and detected  person images. The 20 training/test splits provided  in ~\cite{li2014deepreid} is used under and the single-shot setting as in \cite{liao2015person} -- two images are randomly chosen for testing; one is for probe and the other for gallery.

\noindent{\bf Market1501}~\cite{zheng2015scalable} is the biggest re-id benchmark dataset to date, containing 32,668 detected person bounding boxes of 1,501 identities. Each identity is captured by six cameras at most, and two cameras at least. During testing, for each identity, one query image in each camera is  selected, therefore multiple queries are used for each identity. Note that, the selected 3,368 queries in \cite{zheng2015scalable} are hand-drawn, instead of DPM-detected as in the gallery. Each identity may have multiple images under each camera. 
We use the provided fixed training and test set, under both the single-query and multi-query evaluation settings.

\noindent{\bf Feature Representations} \quad
By default the recently proposed  Local Maximal Occurrence (LOMO) features \cite{liao2015person} are used for person representation. 
The descriptor has 26,960 dimensions. To test our method's ability to fuse different representations, we also consider another histogram-based image descriptor
proposed in \cite{lisanti2014matching}. These include colour histogram, 
HOG  and  LBP  which are concatenated resulting in  5138 dimensions.

\noindent{\bf Evaluation metrics} \quad We use Cumulated Matching Characteristics (CMC) curve to evaluate the performance of person re-identification methods for all datasets in this paper. 
Due to space limitation and for easier comparison with published results, we only report the cumulated matching accuracy at selected ranks in tables rather than plotting the actual curves.
Note that for the Market1501 dataset, since there are on average 14.8 cross-camera ground truth matches for each query, we additionally use mean average precision(mAP) as in \cite{zheng2015scalable} to evaluate the performance.

\noindent{\bf Parameter setting} \quad There is no free parameter to tune for our model. However, with the kernelisation, kernel selection is necessary. Unless stated otherwise, RBF kernel is used with the kernel width determined automatically using the mean pairwise distance of samples. For other compared methods, different model specific parameters have to be tuned carefully to report the highest results. Note that under the semi-supervised null space learning algorithm, there are free parameters: the value of $k$ in the $k$-nn graph is fixed to 3  for all experiments. The percentage of neighbours $f$ kept for creating pseudo classes are fixed at 40\%. We found that the results are not sensitive to the values of these parameters. 

\subsection{Fully Supervised Learning Results}

 For the fully supervised setting, all the labels of the training data are used for model learning. For different datasets, we select different most representative and competitive alternative methods for comparison.

\noindent{\bf Results on VIPeR} \quad We first evaluate our method against the state-of-the-art on VIPeR. We compare with 17 existing methods. Among them, the distance metric learning based methods are  RPLM~\cite{hirzer2012relaxed}, MtMCML~\cite{ma2014person}, Mid-level Filter~\cite{zhao2014learning}, SCNCD~\cite{yang2014salient}, Similarity Learning~\cite{chen2015similarity}, LADF ~\cite{li2013learning},  ITML ~\cite{davis2007information}, LMNN ~\cite{weinberger2005distance}, KISSME ~\cite{koestinger2012large}, and MCML ~\cite{MCML_GlobersonR05}, whilst the others are discriminative subspace learning based methods including kCCA ~\cite{lisanti2014matching}, MFA ~\cite{xiong2014person}, kLFDA \cite{xiong2014person}, and  XQDA ~\cite{liao2015person}. Note that XQDA can be considered as hybrid between metric learning and subspace learning. In addition, deep learning based model is also compared \cite{ahmed2015improved}.  For fair comparison, whenever possible (i.e.~code is available and features can be replaced), we compare with these methods using the same LOMO features. Otherwise, the reported results are presented.

From the results shown in Table \ref{VIPER}, we can make the following observations: (1) Our method achieves the highest performance when a single type of features are used (Rank 1 of 42.28\% compared to the closest competitor XQDA ~\cite{liao2015person} which gives 40.00\%). (2) For fair comparison against methods which fuse more than one types of features \cite{paisitkriangkrailearning} or more than one models \cite{zhao2014learning}, we also present our method's result obtained by a simple score-level fusion using the two types of features described earlier. Our method (Ours (Fusion)) beats the nearest rival \cite{paisitkriangkrailearning} by over 5\% on Rank 1. (3) The discriminative subspace learning based methods seem to be more competitive compared with the distance metric learning based methods. Note that all of them have been kernelised and we observe a significant drop in performance without kernelisation. This confirms the conclusion drawn in \cite{xiong2014person} that kernelisation is critical for addressing the non-linearity problem in re-id.  (4) The most related methods MCML ~\cite{MCML_GlobersonR05} and MtMCML~\cite{ma2014person} yeild much poorer results\footnote{The result of MCML is from \cite{ma2014person} using different features. We did have access to the code of MCML. However, no matter how hard we try, it would not converge to a meaningful solution using the higher-dimensional LOMO features. }, indicating that the principle of collapsing same-class samples is better realised in a subspace learning framework which provides an exact and closed-form solution. (5) The deep learning based method \cite{ahmed2015improved} does not fare well on this small dataset despite the fact that the model has been pre-trained on the far-larger CUHK01+CUHK03 datasets. This suggests that the model learned from other datasets are not transferable by the simple model fine-tuning strategy and small sample size remains a bottle-neck for applying deep learning to re-id.  

		\setlength{\tabcolsep}{5pt}
	\begin{table}
		\centering
	\caption{Fully supervised results on VIPeR}	
		\label{VIPER}
		\begin{tabular}{|l||c c c c|}
		\hline

			{\em Rank} & 1 & 5 & 10 & 20 \\

			\hline
			\hline

RPLM~\cite{hirzer2012relaxed} & 27.00  & 55.30 & 69.00  & 83.00 \\
MtMCML~\cite{ma2014person}& 28.83 & 59.34 & 75.82 & 88.51\\ 
MCML~\cite{MCML_GlobersonR05}&20.19 &47.31  &63.96  &77.69 \\
Mid-level~\cite{zhao2014learning}& 29.11 & 52.34 & 65.95 & 79.87\\ 
SCNCD~\cite{yang2014salient} & 37.80 & 68.50 & 81.20 & 90.40 \\
LADF~\cite{li2013learning}&30.22 &64.70  &78.92   &90.44 \\
Improved Deep~\cite{ahmed2015improved}&34.81 &63.61 &75.63 &84.49 \\
Similarity Learning~\cite{chen2015similarity} &36.80 &70.40 & \bf 83.70 &91.70\\
ITML (LOMO)~\cite{davis2007information}&24.65 &49.78 &63.04 &78.39\\
LMNN (LOMO)~\cite{weinberger2005distance}&29.43 &59.78 &73.51 &84.91\\
KISSME (LOMO)~\cite{koestinger2012large}&34.81 &60.44 &77.22 &86.71\\
kCCA (LOMO)~\cite{lisanti2014matching} & 30.16 & 62.69 & 76.04 & 86.80 \\
MFA (LOMO)~\cite{xiong2014person}&38.67 &69.18 &80.47 &89.02 \\
kLFDA (LOMO)~\cite{xiong2014person}&38.58 &69.15 &80.44 &89.15\\
XQDA (LOMO)~\cite{liao2015person} &40.00 &68.13 &80.51 &91.08 \\
Ours (LOMO) &\bf 42.28 & \bf 71.46 &82.94 &\bf 92.06 \\
\hline
\hline
Mid-level+LADF~\cite{zhao2014learning} & 43.39 & 73.04 & 84.87 & 93.70 \\
Metric Ensembles~\cite{paisitkriangkrailearning} & 45.90  &77.50  &88.90 &95.80    \\
 Ours (Fusion)& \bf 51.17 & \bf 82.09 & \bf 90.51 & \bf 95.92 \\
\hline

		\end{tabular}
	\end{table}

\noindent{\bf Results on PRID2011} \quad  We compare the state-of-the-art~\cite{hirzer2012relaxed, paisitkriangkrailearning} results reported on PRID2011 in Table \ref{prid2011}. With access to the implementation codes, we also compare with the methods in~\cite{liao2015person, xiong2014person, lisanti2014matching} using the same LOMO features. The results show clearly with a single feature type, our method is the state-of-the-art; when fusing two types of features, the result is improved dramatically (over 10\% increase on both Rank 1 and 5), and significantly higher than the reported results of the feature fusion method in \cite{paisitkriangkrailearning}, which fuses four different types of features including the deep convolutional neural network (CNN) features.

	\begin{table}[H]
		\centering
		\caption{Fully supervised results on PRID2011}
		\label{prid2011}
		\scalebox{0.9}{
		\begin{tabular}{|l||c c c c|}

			\hline

			{\em Rank} & 1 & 5 & 10 & 20 \\

			\hline
			\hline
RPLM~\cite{hirzer2012relaxed} & 15.00  & 32.00 & 42.00  & 54.00  \\
kCCA (LOMO)~\cite{lisanti2014matching} & 14.30 & 37.40 & 47.60 & 62.50 \\
MFA (LOMO)~\cite{xiong2014person}&22.30 &45.60 &57.20 &68.20 \\
kLFDA (LOMO)~\cite{xiong2014person}&22.40 &46.50 &58.10 &68.60\\
XQDA (LOMO)~\cite{liao2015person}&26.70 &49.90 &61.90 &73.80\\
 Ours (LOMO) & \bf 29.80 & \bf  52.90 & \bf  66.00 & \bf 76.50   \\
\hline
\hline
Metric Ensembles~\cite{paisitkriangkrailearning} & 17.90 &39.00 &  50.00  &62.00 \\
 Ours (Fusion)& \bf 40.90 & \bf 64.70 & \bf 73.20 & \bf 81.00 \\
\hline

		\end{tabular}
}
	\end{table}

\noindent{\bf Results on CUHK01 \& CUHK03} \quad Compared with VIPeR and PRID2011, these two datasets are much bigger with thousands of training samples. However, the sample size is still much smaller than the feature dimension, i.e.~the SSS problem still exists. Table \ref{cuhk01}  shows that on CUHK01, our method beats all compared existing methods at low ranks and when two types of features are fused, the margin is significant. As for CUHK03, there are two versions: the one with manually cropped person images, and the one with bounding boxes produced by a detector. The latter obviously is harder  as reflected by the decrease of matching accuracy for all compared methods. But it is also a better indicator of real-world performance. It can be seen from Table \ref{cuhk03} that, as expected, on this much larger dataset, the deep learning based model \cite{ahmed2015improved} with its millions of parameters becomes much more competitive -- with manually cropped images, our result with single feature type is higher on Rank 1 but lower on other ranks. However, with the detector boxes, our method is less affected and outperforms the deep model in \cite{ahmed2015improved} by a big margin. In addition, our performance is further boosted by fusing two types of features.

	\begin{table}[H]
		\centering
		\caption{Fully supervised results on CUHK01}
		\label{cuhk01}
		\scalebox{0.9}{
		\begin{tabular}{|l||c c c c|}
		\hline

		{\em Rank} & 1 & 5 & 10 & 20 \\

		\hline
		\hline

SalMatch~\cite{zhao2013person} & 28.45 &45.85  & 55.67 & 67.95 \\
Mid-level Filter~\cite{zhao2014learning} & 34.30 &55.06  & 64.96 & 74.94 \\
Improved Deep~\cite{ahmed2015improved}&47.53 &71.60 &80.25 &87.45 \\
kCCA (LOMO)~\cite{lisanti2014matching} &56.30 &80.66  &87.94 &93.00 \\
MFA (LOMO)~\cite{xiong2014person}&54.79 &80.08 &87.26 &92.72 \\
kFLDA (LOMO)~\cite{xiong2014person} &54.63 &80.45  &86.87 &92.02 \\
XQDA (LOMO)~\cite{liao2015person} & 63.21 & 83.89  &\bf 90.04 & 94.16 \\
Ours (LOMO)&\bf 64.98 &\bf 84.96 & 89.92 &\bf 94.36 \\
\hline
\hline
Metric Ensembles ~\cite{paisitkriangkrailearning} & 53.40 &76.40 &84.40 & 90.50 \\
 Ours (Fusion) &\bf 69.09 &\bf 86.87 & \bf 91.77 &\bf 95.39 \\
\hline

		\end{tabular}
}
	\end{table}

			\setlength{\tabcolsep}{2.5pt}
			\renewcommand{\arraystretch}{1.2}
	\begin{table}
		\centering
		\caption{Fully supervised results on CUHK03. '-' means that no reported results is available.}
		\label{cuhk03}
		\footnotesize
		\scalebox{0.9}{
		\begin{tabular}{|l||c c c c|c c c c|}
		\hline
		{\em Dataset}  & \multicolumn{4}{c|}{CUHK03 (Manual)} & \multicolumn{4}{c|}{CUHK03 (Detected)}\\

	    \hline

		{\em Rank} & 1 & 5 & 10 & 20 & 1 & 5 & 10 & 20\\

		\hline
		\hline

DeepReID~\cite{li2014deepreid}& 20.65 &51.50  &66.50  &80.00  & 19.89 &50.00  &64.00  & 78.50\\
Improved Deep~\cite{ahmed2015improved}&54.74 & \bf 86.50 &\bf 93.88 & \bf 98.10 &44.96 &76.01 &83.47 &93.15\\
XQDA (LOMO)~\cite{liao2015person} & 52.20  &82.23 &92.14 &96.25 & 46.25&78.90 &88.55 &94.25\\
Ours (LOMO)&  \bf 58.90 &85.60 &92.45 &96.30 &\bf 53.70 &\bf 83.05 &\bf 93.00 &\bf 94.80 \\
\hline
\hline
Metric Ensembles ~\cite{paisitkriangkrailearning}  & 62.10 & 89.10  &94.30  & 97.80   & - & - & - &- \\
 Ours (Fusion) &\bf 62.55  &\bf 90.05 &\bf 94.80 &\bf 98.10 &\bf 54.70 &\bf 84.75 &\bf 94.80 &\bf 95.20 \\
\hline

		\end{tabular}
		}
	\end{table}

\noindent{\bf Results on Market1501} \quad This dataset is the largest and most realistic dataset with natural detector errors abundant in the provided data as they were collected in front of a busy supermarket. Since it is new, few reported results are available. The baseline presented in \cite{zheng2015scalable} is not competitive because it is based on a weaker BoW features and L2-Norm distance. We compare our method with four alternatives with the same LOMO features. The results in  Table \ref{market} again show that our method significantly outperforms the alternatives, under both the single query and multi-query settings and with both evaluation metrics.  This is despite the fact that with 12,936 training samples, the SSS problem is the least severe in this dataset.

	\begin{table}[H]
		\centering
		\caption{Fully supervised results on Market1501}
		\label{market}
		\scalebox{0.9}{
		\begin{tabular}{|l||c c| c c |}
	    \hline

		{\em Query} &\multicolumn{2}{c|}{\em singleQ} &\multicolumn{2}{c|}{\em multiQ} \\
		\hline
		{Evaluation metrics} &{Rank-1} &{mAP} &{Rank-1} &{mAP}\\

		\hline
		\hline

Baseline~\cite{zheng2015scalable}&34.38 &14.10 &42.64 &19.47 \\
Baseline (+HS)~\cite{zheng2015scalable}&- &- &47.25 &21.88 \\
\hline
\hline
KISSME (LOMO)~\cite{koestinger2012large}&40.50 &19.02 & - & -\\
MFA (LOMO)~\cite{xiong2014person}&45.67 &18.24 &- &- \\
kLFDA (LOMO)~\cite{xiong2014person}&51.37 &24.43 &52.67 &27.36 \\
XQDA (LOMO)~\cite{liao2015person}&43.79 &22.22 &54.13 &28.41 \\
Ours (LOMO)& \bf 55.43 &\bf 29.87 & \bf 67.96  & \bf 41.89 \\
\hline
\hline
Ours (Fusion)& \bf 61.02 &\bf 35.68 & \bf 71.56  & \bf 46.03 \\
\hline

		\end{tabular}
		}
	\end{table}

\subsection{Semi-supervised Learning Results}

For semi-supervised setting, we use the VIPeR and PRID2011 datasets. The same data splits are used  as in the fully-supervised setting. The difference is that only one third of  the training data are labelled following the setting  in~\cite{liu2014semi, kodirovdictionary}. For comparison, apart from the state-of-the-art methods in \cite{liu2014semi, kodirovdictionary}, we also choose three subspace learning based methods trained on the labelled data only. 

The results in Table \ref{table:semi-supervised} show that the performance of our method is clearly superior to that of the compared alternatives. The advantage is more significant on PRID2011. This dataset has only 100 pairs or 200 training samples; with only one third of them labelled, the SSS problem becomes the most acute than any experiment we conducted before. Comparing Table \ref{table:semi-supervised} with Table \ref{prid2011}, it is apparent that the performance of all three compared subspace learning methods, kCCA, kLFDA, and XQDA degrades drastically. In contrast, the performance of our method decrease much more gracefully from 29.80\% to 24.70\% on Rank 1. This is partly because our self-training based method can exploit the unlabelled data. It also shows that it can better cope with the SSS problem in its extreme.
    
		\setlength{\tabcolsep}{2.5pt}
		\renewcommand{\arraystretch}{1.25}
	\begin{table}[H]
		\centering
		\caption{Semi-supervised Re-ID results on VIPeR and PRID2011}
		\label{table:semi-supervised}
		\footnotesize
		\scalebox{0.85}{
		\begin{tabular}{|l||c c c c|c c c c|}
		\hline
		{\em Dataset} &  \multicolumn{4}{c|}{VIPeR} & \multicolumn{4}{c|}{PRID2011}\\

	    \hline

		{\em Rank} & 1 & 5 & 10 & 20 & 1 & 5 & 10 & 20\\

		\hline
		\hline
SSCDL~\cite{liu2014semi}&25.60 &53.70 &68.20 &83.60 &- &- &- &- \\
kCCA (LOMO)~\cite{lisanti2014matching}&13.64 &37.97 &53.77 &69.94 &5.80 &16.00  &24.70  &36.00  \\
kLFDA (LOMO)~\cite{xiong2014person}&25.47 &53.25 &66.49 &80.13 &12.00  &27.10  &37.80  &50.30  \\
XQDA (LOMO)~\cite{liao2015person}&28.04 &56.30 &69.65 &81.74 &12.60 &29.40 &40.20 &53.00 \\
IterativeLap (LOMO)~\cite{kodirovdictionary}&29.43  &49.05  &59.18  & 69.62 &18.70  &34.60  &43.50  &52.30  \\
\hline
\hline
Ours (LOMO)&\bf 31.68 &\bf 59.40 &\bf 72.78 &\bf 84.91 &\bf 24.70 &\bf 46.80 &\bf 58.20 &\bf 68.20 \\
\hline
\hline
Ours (Fusion)&\bf 41.01  &\bf 69.81 &\bf 81.61 &\bf 91.04 &\bf 35.80 &\bf 58.10  &\bf 69.10 &\bf 78.90 \\

\hline

		\end{tabular}
		
		}
	\end{table}

\subsection{Running Cost}

%

We compare the run time of our method with XQDA, kLFDA and MFA on  Market1501. We calculate the overall training time over 12,936 samples and test time over 3,368 queries. All algorithms are implemented in Matlab and run on a server with 2.6GHz CPU cores and 384GB memory. Table \ref{runningtime} shows that for training, our method is the most efficiently, whilst on testing it is much slower than XQDA, but faster than kLFDA and MFA. Considering the test time is over 3,368 queries, it is more than adequate for real-time applications.

			\setlength{\tabcolsep}{3pt}
	\begin{table}[H]
		\centering
		\caption{Run time comparison on Market1501 (in seconds)}
		\label{runningtime}
		\scalebox{0.9}{
		\begin{tabular}{|l||c| c| c| c |}
		\hline

			{\em Method} &\ Ours & XQDA~\cite{liao2015person} & kLFDA~\cite{xiong2014person} & MFA~\cite{xiong2014person} \\

			\hline
			\hline
Training&393.1  &3233.8  &995.2 &437.8  \\
Testing& 31.3  &1.6 &43.4  &43.2   \\
\hline

		\end{tabular}
		}
	\end{table}

\section{Conclusion}
We proposed to solve the person re-id problem by learning a discriminative null space of the training samples. Compared with existing re-id models, the employed NFST model is much simpler, with a closed-form solution and no parameters to tune. Yet, it is very effective in dealing with the SSS problem faced by the re-id methods. Extensive experiments on five benchmarks show that our method achieves the state-of-the-art performance on all of them under both fully supervised and semi-supervised settings. 

\section*{Acknowledgement}
This work was funded in part by the European FP7 Project SUNNY (grant agreement no. 313243).

{\small
\bibliographystyle{ieee}
\bibliography{egbib,matching_partial}
}

\end{document}